\documentclass[sn-mathphys,Numbered]{sn-jnl}% Math and Physical Sciences Reference Style
\usepackage{graphicx}%
\usepackage{multirow}%
\usepackage{amsmath,amssymb,amsfonts}%
\usepackage{amsthm}%
\usepackage{mathrsfs}%
\usepackage[title]{appendix}%
\usepackage{xcolor}%
\usepackage{textcomp}%
\usepackage{manyfoot}%
\usepackage{booktabs}%
\usepackage{algorithm}%
\usepackage{algorithmicx}%
\usepackage{algpseudocode}%
\usepackage{listings}%
\usepackage{graphicx}
\usepackage{tikz}
\usepackage{lingmacros}
\usepackage{tree-dvips}
\usepackage{pgfplots}
\usepackage{xlop}
\pgfplotsset{width=7cm,compat=1.9}
\usepackage{scalerel}
\theoremstyle{thmstyleone}%
\theoremstyle{thmstyletwo}%

\theoremstyle{thmstylethree}%

\begin{document}

\title[Article Title]{Automated Discovery of Integral with Deep Learning}


\author*[1]{\fnm{Xiaoxin} \sur{Yin}}\email{xiaoxin@gmail.com}

\affil*[1]{\orgname{Independent Researcher}, \orgaddress{\city{San Jose}, \state{California}, \country{USA}}}


\abstract{Recent advancements in the realm of deep learning, particularly in the development of large language models (LLMs), have demonstrated AI's ability to tackle complex mathematical problems or solving programming challenges. However, the capability to solve well-defined problems based on extensive training data differs significantly from the nuanced process of making scientific discoveries. Trained on almost all human knowledge available, today's sophisticated LLMs basically learn to predict sequences of tokens.  They generate mathematical derivations and write code in a similar way as writing an essay, and do not have the ability to pioneer scientific discoveries in the manner a human scientist would do.
	
In this study we delve into the potential of using deep learning to rediscover a fundamental mathematical concept: integrals. By defining integrals as area under the curve, we illustrate how AI can deduce the integral of a given function, exemplified by inferring $\int_{0}^{x} t^2 dt = \frac{x^3}{3}$ and $\int_{0}^{x} ae^{bt} dt = \frac{a}{b} e^{bx} - \frac{a}{b}$. Our experiments show that deep learning models can approach the task of inferring integrals either through a sequence-to-sequence model, akin to language translation, or by uncovering the rudimentary principles of integration, such as $\int_{0}^{x} t^n dt = \frac{x^{n+1}}{n+1}$.}




\maketitle

\section{Introduction}\label{intro}

The recent advances in deep learning, especially those in large language models (aka LLM), have shown the possibility of an AI agent performing any task a human can perform, including scientific research. Recent studies have shown that LLMs such as GPT-4\cite{gpt-4}, Microsoft Copilot\cite{ms-copilot}, and CodeLlama\cite{roziere23} can solve competition-level coding problems \cite{huang23}, and LLMs such as GPT-4 and Llemma\cite{azerbayev23} can solve some high-school-level competition math problems (including some IMO-level problems). These LLMs can certainly help researchers solve some problems they encounter in their daily research. 

However, being able to solve a type of well-defined problems is very different from making discoveries in scientific research. For instance, in order to train an LLM to solve coding problems, a general-purpose LLM is often fine-tuned on all public code on GitHub, and also fine-tuned on hundreds of thousands of coding problems from various platforms such as CodeForce and LeetCode. For example, CodeLlama-Python  underwent fine-tuning with 100 billion tokens of Python code. The LLM simply learns how to write code given the coding problem (which is the prompt), by learning to predict the next token in its code given the prompt and tokens it has generated. This is essentially the same methodology used to train a model to write novels after reading millions of novels. It does not have the capability of discovering what is not already known to it, making it unable to make scientific discoveries like a scientist would do.

Very recently DeepMind published their pioneer work on training a deep learning model to solve IMO level geometry problems, without using theorems found by human \cite{trinh24}. The authors randomly generated geometric data and let the model learn the geometric rules from one billion examples. This is an exciting piece of work in which an AI discovers mathematical theorems and methodologies on its own.

In this paper we would like to explore how to use deep learning models to discover an old but most important tool: Integrals. We define integral of a function $f$ using area under the curve from 0 to $x$ (i.e., the integral of a function $f(x)$ is defined as  $\int_{0}^{x} f(t) dt$), which can be easily calculated by sampling a function (e.g., using the simple Trapezoidal rule). We hope to answer the following question: Simply given this definition, can deep learning models and algorithms learn to derive the mathematical expression of the integral of a function, and find the basic rules of integrals for each type of functions (polynomials, exponential, trigonometry, etc.)

In this study the AI models are only given the following data and tools: 
\begin{itemize}
	\item A large set of randomly generated univariate functions.
	\item The ability to evaluate a function.
	\item The definition of integral (defined as area under the curve), and a program that can calculate the area under the curve of any function using the Trapezoidal rule, which mainly calculates the cumulative sum of a function. 
\end{itemize}

Our experiment shows that given the above data and tools, AI models can successfully finish the following two tasks:
\begin{itemize}
  \item Based on the mathematical expression of a function $f(x)$, the model can directly infer the mathematical expression of the integral of $f(x)$, without evaluating the function. The process is actually similar to translating a sentence from one language to another one.
  \item Discover the basic rules for integrating each simple type of functions, such as polynomials, trigonometric, and exponential functions. 
\end{itemize}

This experiment shows that AI models can acquire the capability of integration, without human providing any examples. The only information required is how integral is defined (as area under the curve in our study), and AI can derive the rest. Please note that we train our AI models from scratch and it never sees any information other than what we provide in our experiment.

Our experiment is consisted of four steps. First we compute the definite integral ($\int_{0}^{x} f(t) dt$) of each random function $f$. Then we use a symbolic regression model\cite{dAscoli22} to infer the mathematical expression of the definite integral $g(x) = \int_{0}^{x} f(t) dt$. Please note the symbolic regression model was trained only on randomly generated functions and their curves, and was not aware of the concept of integral. The third step is to train a language model to infer the integral function $g$ from $f$ without calculating area under the curve, just like what a college student would do in the calculus class. Lastly, we try to infer rules of integral for each simple type of functions. For example, $\int_{0}^{x} ae^{bt} dt = \frac{a}{b} e^{bx} - \frac{a}{b}$, where $a$ and $b$ are constants.

In this process we do not utilize any LLMs trained on human knowledge, because they probably have learned about integrals during training. The only training data we use are randomly generated functions, their curves, and their areas under the curves. 

We have no doubt that an AI agent would be more powerful if it has been trained on human knowledge and thus is capable to perform reasoning like a human. But that also means the agent has read all human knowledge, and we can only test its discovery power on what humans have not discovered yet, which is certainly a highly challenging problem. Therefore, in this study we choose to study how AI can discover an important math concept like integral, without using any prior human knowledge. This is the start of our research and we will expand to more complex problems in near future.

\section{Related Work}\label{relwork}

The idea of automating scientific research activities dates back to the early days of computer science. An article on \emph{Science} in 2009 \cite{waltz09} provides a great overview on the early explorations. Also in 2009 a ``Robot Scientist'' named Adam was released \cite{king09}. The authors developed specialized hardware for conducting basic experiments, such as tracking yeast growth with varying gene deletions and metabolites. This was paired with logic programming software for selecting experiments. The software keeps track of various hypotheses and chooses experiments likely to refute many of them at once. These experiments are automatically performed, and their results guide the next experiment's selection. Adam effectively identified the functions of multiple genes, requiring fewer experiments compared to other experiment-selection methods like cost-based choices. \cite{naik16} presents a research that utilizes special hardwares to automatically learn the effects of different drugs upon the distribution of different proteins within mammalian cells.

Very recently a breakthrough was brought by DeepMind \cite{trinh24}, in which the authors created a large language model that learned geometry on one billion generated problems, in order to discover geometry properties, and train itself to prove these properties. The model was tested on 30 IMO geometry and got 25 of them correct, which outperforms the majority of IMO participants. This is the first time a neutral network model learns to master a discipline of science on its own, and it will not be surprising if the same methodology can be extended to other disciplines such as number theory and combinatorics. 

Our goal is to let AI make scientific discoveries on its own, and we start from a simple but important problem raised hundreds of years ago: Integrals. We are following a similar methodology as  \cite{trinh24} to let a model learn how to compute integrals of functions on its own. 

On the other hand, there have been studies on training deep learning models to infer the integrals of functions. In \cite{lample19} the authors treated the problem of inferring the integral of a function as a translation problem, and trained a traditional transformer\cite{vaswani17} using pairs of randomly generated functions and their integrals. This is very different from our study, as it was intended to teach a model how to infer the integral of a function by providing many examples, while we are trying to let models discover how to compute integrals on its own.

\section{Experiment Design and Model Training}\label{model}

As mentioned in Section \ref{intro}, our experiment is consisted of four steps:

\begin{itemize}
\item Collect a large set of functions (including randomly generated ones), and for each function $f$, compute the integral defined as area under the curve (i.e., $\int_{0}^{x} f(t) dt$). 
\item For each function $f$, use a symbolic regression model to infer the function for its integral: $g(x) = \int_{0}^{x} f(t) dt$. For example, if $f(x) = x^2$, then ideally the symbolic regression model should infer that $g(x) = \frac{x^3}{3}$.
\item Train a language model which directly infers the mathematical expression of $g(x)$ from $f(x)$.
\item Try to identify simple rules of integral for each type of functions, such as $\int_{0}^{x} t^n dt = \frac{x^{n+1}}{n+1}$.
\end{itemize}

We will describe each step in details in the following subsections.

\subsection{Dataset of Functions}\label{randomfunc}

In this study we consider functions that are polynomial, trigonometric, or exponential. 

For polynomial functions, we get 10K of them from the \emph{parametric functions} data in AMPS dataset\cite{hendrycks21} (under folder \emph{mathematica/algebra}) that can be parsed by SymPy\cite{sympy}. Then we randomize each function by randomly increase, decrease, multiply or divide each constant (including coefficient and exponent) by an integer between 1 and 3, without changing its sign. By combining the original functions and the randomized functions, we get a set of 58K polynomial functions.

Then we randomly generate trigonometric or exponential functions such as $a \cdot \sin(bx) + c$ or $a \cdot e^{bx} + c$, where $a$, $b$ and $c$ are random constants between -10 and 10. 114K functions are generated in this way.

Obviously we could have included more types of functions. But we feel the above set is sufficient for studying how deep learning can discover how to calculate integrals.

For each function $f$, we compute the integral defined as area under the curve $\int_{0}^{x} f(t) dt$. This is done using the Trapezoidal rule (by numpy.trapz), which is simply a cumulative sum of a function over a certain range on the $x$-axis.

\subsection{Symbolic Regression}\label{symbolic}

Symbolic regression is the procedure of inferring the mathematical expression of a function given a sample of its values. Traditionally the standard approach are evolutionary programs such as genetic programming  \cite{augusto00}\cite{schmidt09}\cite{murari14}\cite{mckay95}, despite their high computational cost and limited performance. In recent years it has been found that deep neural networks outperforms the traditional approaches \cite{dAscoli22}\cite{sahoo18}\cite{kim20}\cite{petersen19}.

Here we take the approach in \cite{dAscoli22} as an example. It converts a symbolic regression problem as a sequence-to-sequence learning problem. The input to the problem is a sample of points of the target function: $(\vec{x_1}, y_1), ..., (\vec{x_n}, y_n)$. Each dimension of each input point is converted into a few tokens, representing the index of the dimension and the value. The output, which is the mathematical expression of the function, is encoded as a sequence of tokens, each being a variable, operator, function, or value. Then the symbolic regression problem is converted into a sequence-to-sequence prediction problem, which tries to infer the output sequence from the input sequence. The authors adopted a classic transformer architecture \cite{vaswani17}, with an encoder and an decoder, each having 8-layers. Because the transformer is a language model and is not capable of computing function values, the constants in the output mathematical expression can deviate from the actual values. Therefore, the last step of the approach is to use gradient descent to optimize the constants in the output expression, until it reaches a local minimum. There is a certain degree of randomization in the process of inferring the output expression, i.e., generating each token of the output sequence, and the degree of randomness can be controlled by the \emph{temperature} parameter.

In order to train the model for predicting the output sequence given the input sequence, hundreds of thousands of randomly generated functions were used in the training process. The authors reported reasonable high accuracy ($>80\%$) for symbolic regression problems with mathematical expressions of limited complexity.

After comparing various approaches presented in  \cite{landajuela2022} and that in \cite{dAscoli22}, we found that the approach in \cite{dAscoli22} has non-inferior accuracy but is much more time efficient than the others, with a stable latency of about one second. This is probably because it mainly involves a sequence-to-sequence inference using a transformer model.  In contrast, the latency of all other approaches we tested varies greatly from case to case, from a split second to minutes. Since we need to run symbolic regression on hundreds of thousands of cases, a low and stable latency is key to our study, and thus we choose the approach in \cite{dAscoli22}.

\subsection{Model Training}\label{model}

After inferring the mathematically expression of the integral of each function $f$ (defined as area under the curve $g(x) = \int_{0}^{x} f(t) dt$), we train a model that directly infers the mathematical expresion of $g$ from that of $f$. For example, given input $4x^2 + 2$, the model should output $1.33 x^3 + 2 x$. 

Please note that we use floating point number $1.33$ instead of $\frac{4}{3}$ for the following reason. A symbolic regression model usually outputs floating point numbers instead of fractions as coefficients, because it first determines the structure of the mathematical expression with approximate coefficients, and then uses gradient descent or genetic algorithm to find optimal values for the coefficients. Therefore we have to use floating point numbers instead of fractions.

When training a language model to infer the integral of a function, we need to convert the mathematical expressions into sequences of tokens, and thus we need to limit the number of digits to prevent a number from being converted into too many tokens. For example, if there is a coefficient ``0.1428571428571'', it will be tokenized into the following tokens by GPT-Neo \cite{gpt-neo}: ``0'', ``.'', ``14'', ``28'', ``57'', ``14'', ``28'', ``571''. In this study we choose to retain two decimal places, which avoids the complexity introduced by such lengthy representation of numbers. Our experiments show that keeping two decimal places does not hurt its capability to infer the basic rules of integrals.

We test two most popular open-source model architectures. The first one is GPT-Neo\cite{gpt-neo}\footnote{\url{https://github.com/EleutherAI/gpt-neo}} , which is a most popular open-source implementation of the GPT-3 model (with much smaller model size). The second is Google's Flan T5 \cite{flan-t5}, a most popular open-source sequence-to-sequence model.

\begin{figure}[h]%
	\centering
	\includegraphics[width=1.0\textwidth]{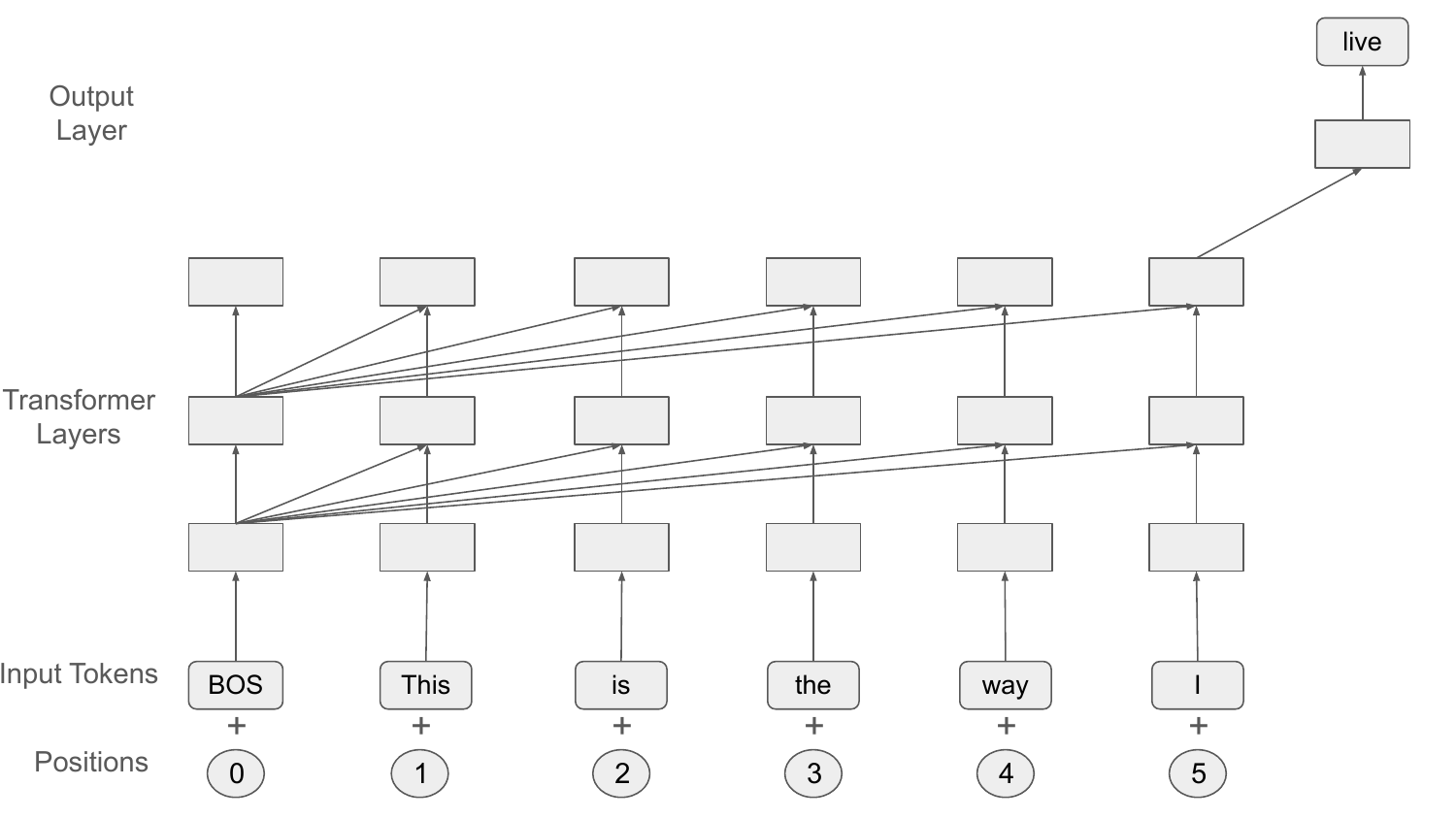}
	\caption{An illustration of how an autoregressive model like GPT works. It predicts each token based on all previous tokens (including a Begin-Of-Sentence token at the beginning). }\label{fig:gpt}
\end{figure}

GPT-Neo is an autoregressive model, which simply predicts the next token based on all previous tokens. Figure \ref{fig:gpt} illustrates how it works. In the transformer architecture, each input token is initially converted into an embedding, representing the token in a numerical form. This token embedding is then combined with a positional embedding, which encodes the position of the token in the sequence. The resultant combined embedding serves as the input to the first layer of the transformer.

Within each layer of the transformer, a core component is the Multihead Self-Attention (MSA) mechanism, which operates on the embeddings (or hidden states from previous layers) by computing interactions between each pair of positions in the sequence. Essentially, it allows each position to consider information from all other positions, facilitating a comprehensive understanding of the entire sequence.

However, to prevent the model from using future information during training (a process known as "looking ahead"), a triangular mask is applied in the self-attention mechanism. This mask ensures that the computation for a given position can only utilize information from preceding positions, preserving the sequential nature of the data.

Finally, at the output stage of the transformer, it's the hidden state corresponding to the last position that is  used for predicting the next token in the sequence. This approach allows the model to generate predictions based on the entire context provided by the input sequence up to that point.

\begin{figure}[h]%
	\centering
	\includegraphics[width=1.0\textwidth]{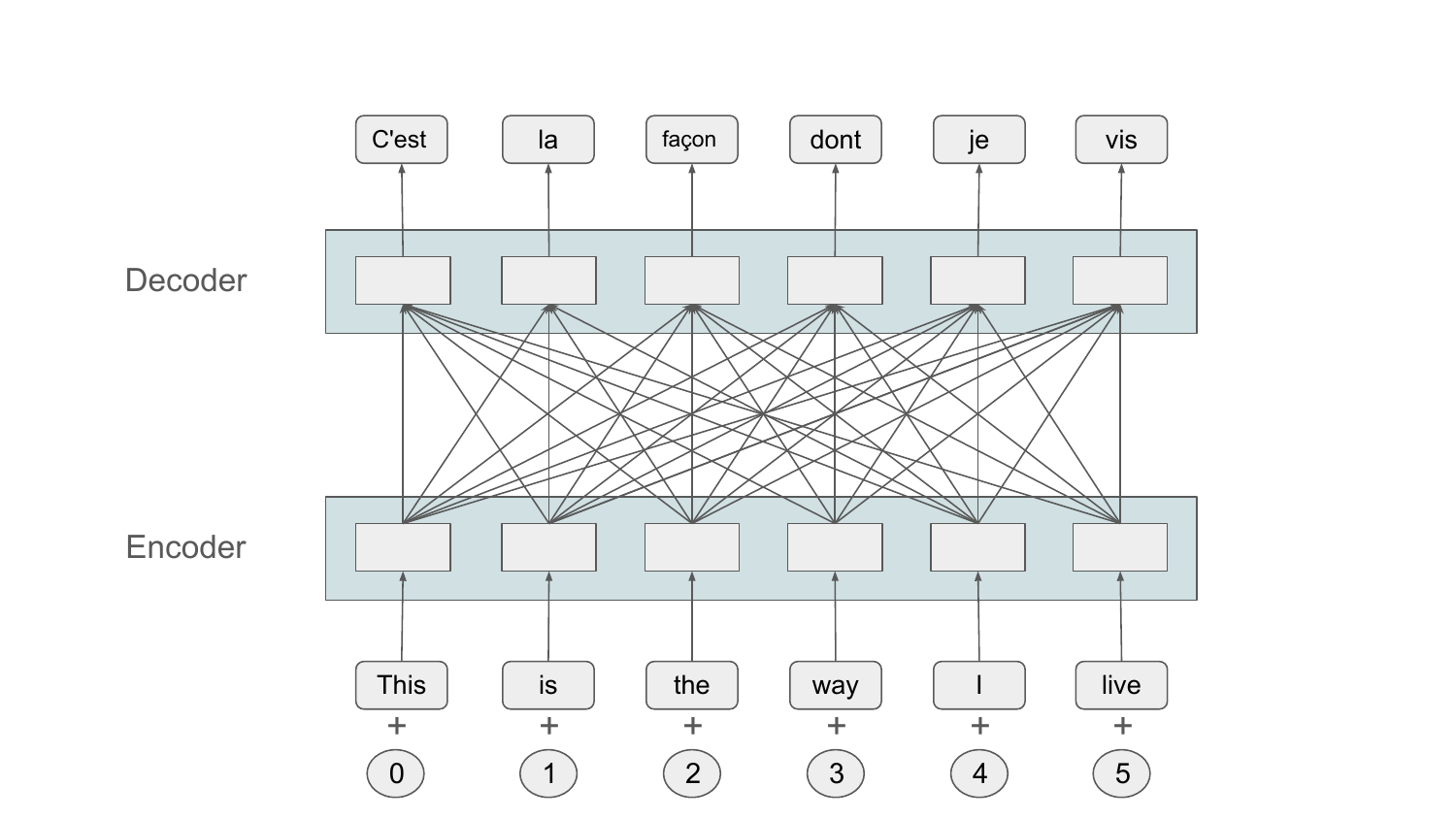}
	\caption{An illustration of a sequence-to-sequence prediction, which translates an English sentence into French. }\label{fig:seq2seq}
\end{figure}

Flan T5 stands out as a sequence-to-sequence model, as illustrated in Figure \ref{fig:seq2seq}. This model operates by first processing a sequence of tokens, transforming them into embeddings which are then input into an encoder. This encoder is essentially a stack of transformer layers, typically ranging from 12 to 48 in number. In the encoder's Multihead Self-Attention component within each layer, there is no triangular mask. Consequently, each input token has the potential to impact the hidden state at any position across the sequence.

Parallel to the encoder, the decoder in Flan T5 is also based on the transformer architecture, generally mirroring the encoder in terms of the number of layers. In each layer of the decoder, the Multihead Self-Attention mechanism functions to calculate interactions between the hidden states of the encoder and those of the decoder. This process is critical for aligning and translating the sequence from the input to the output.

The decoder operates in an autoregressive manner, similar to the mechanism depicted in Figure \ref{fig:gpt}. It sequentially predicts each next token, relying on the collective context provided by all previously predicted tokens. This approach allows a sequence-to-sequence model to generate output sequences that are coherent  based on the input sequence processed by the encoder.

We tried both GPT-Neo and Flan-T5 to train models to predict the integral of a function. Given a dataset of more than 100K functions as described in Section \ref{randomfunc}, for each function $f$, we compute its integral (defined as area under the curve) $g(x) = \int_{0}^{x} f(t) dt$ using the Trapezoidal rule. Please note that so far we only know the function values of $g(x)$ over a range on the $x$-axis, not the mathematical expression. Then we use a symbolic regression model to fit the above function values, which leads to a function $g'(x)$. The symbolic regression model is not always accurate, and we ignore a case if 

\begin{equation}\label{symreg_diff}
	\frac{\int_{0}^{T} |g(x) - g'(x)| dx}{T \cdot max(max(|g(x)|), max(|g'(x)|))} > \epsilon
\end{equation}

In equation \ref{symreg_diff}, $[0, T]$ is the range of our integral, and we calculate average relative difference between $g(x)$ and $g'(x)$, which is the average absolute value of the gap between $g(x)$ and $g'(x)$ divided by the max absolute value of either function. If this average relative difference is larger than our threshold $\epsilon$, we ignore this case because the symbolic regression has probably been incorrect. Since we only have the values of $g(x)$ on a large sample of points, as computed using the Trapezoidal rule, we simply compute the average of $ |g(x) - g'(x)|$ over these points. 

For all the remaining functions with integral functions $g'(x)$ inferred, we consider the original $f$ and the inferred $g'(x)$ as a training example. All the training examples are split into training, validation, and testing set. A model is trained on the training set, and we tune the hyper parameters of the model, such as learning rate and number of epochs trained, in order to minimize the loss on the validation set. Then we test the model on the testing set and report the performance.

\subsection{Discovering Rules of Integral}\label{discover-rule}

Every one with knowledge of calculus knows the basic rules of integral, such as $\int x^n dx = \frac{1}{n+1}x^{n+1} + C$, and $\int sin(x) dx = -cos(x) + C$. In this study we try to design a way so that these rules can be discovered automatically. Consider the deep learning models described in Section \ref{model}. They can infer the integral function from the original function (e.g., output $0.2 x^{5}$ given input $f(x)=x^4$), but cannot output the more generic rules of integrals in a human-understandable manner. In this section we will discuss how to produce these basic rules explicitly.

Here we only aim at finding very simple rules. Take polynomials as an example. If $f(x)$ is a polynomial function, each coefficient of the integral function $g(x)$ can be inferred from one or a few coefficients of $f(x)$ combined with basic arithmetic operations. For example, if we consider quadratic functions, $f(x)=a_2 x^2 +a_1 x + a_0$, and its integral $g(x) = a_3' x^3 + a_2' x^2 + a_1' x + a_0$, the basic rules would be $a_3' =\frac{a_2}{3}$, $a_2' = \frac{a_1}{2}x^2$, $a_1' = a_0$, and $a_0'=0$\footnote{Since we define integrals as area under the curve starting from $x=0$, $g(0)$ is always zero.}. We can also go one step further to produce the rule for all polynomials: $a_n' = \frac{1}{n} a_{n-1}$. 
	
We design a \emph{rule search system} that automatically searches for such simple rules that determines each coefficient in the integral function $g(x)$. Given a large set of functions of the same form (e.g., $a_4 sin(a_3 x) + a_0$) It first selects the most common form of their integral functions ($a_4' cos(a_3' x) + a_1' x + a_0'$), and considers that to the the form of integral function $g(x)$. For each target coefficient (i.e., coefficient in $g(x)$), it first runs a linear regression between the target coefficient $a_k'$ and each coefficient $a_i$ in $f(x)$ (also with $\frac{1}{a_i}$). If any of the linear regressions has $|R^2| > 0.95$, it considers that to be the formula for this target coefficient. If no such linear regression is found, the \emph{rule search system} tries every pair of coefficients $a_i$ and $a_j$, combined with every possible arithmetic operator (such as $a_i \cdot a_j$, $\frac{a_i}{a_j}$, or $a_i + c \cdot a_j$ where $c$ is any constant). If a linear regression between the target coefficient and the pair of coefficients combined has $|R^2| > 0.95$, it stops searching and declares success. Otherwise it keeps searching for triples of coefficients, and stops there (either succeeds or fails).

Our experiments show that the above system successfully discovers formula for integrating functions in our dataset. For example, it discovers that for $f(x) = a_4 exp(a_3x) +a_0$, its integral is $\int_{0}^{x} f(t) dt = \frac{a_4}{a_3} exp(a_3x) + a_0x - \frac{a_4}{a_3} $. Similarly it discovers that for $f(x) = a_4 sin(a_3x) +a_0$, its integral is $\int_{0}^{x} f(t) dt = -\frac{a_4}{a_3} cos(a_3x) + a_0x + \frac{a_4}{a_3} $.

For polynomials, it discovers 

\begin{equation*}
	\begin{split}
		\int_{0}^{x} a_0 + a_1t+a_2t^2 + a_3t^3 + a_4t^4 +a_5t^5 + a_6t^6 dt = \\ a_0t + 0.5 a_1t^2 + 0.333 a_2t^3 + 0.25 a_3t^4 + 0.2 a_4t^5 + 0.167 a_5t^6 + 0.143 a_6t^7 
	\end{split}
\end{equation*}

It also discovers that the new coefficient for $k$-th power $a_k' = \frac{a_{k-1}}{k}$.

\section{Experiment Results}\label{exp}

 We use the GPT-Neo model\cite{gpt-neo} of 350M trainable parameters, downloaded from Huggingface\footnote{\url{https://huggingface.co/xhyi/PT_GPTNEO350_ATG}}, and the Flan-T5-large model \cite{flan-t5}, which has 783M parameters, also downloaded from Huggingface\footnote{\url{https://huggingface.co/google/flan-t5-base}}. All experiments are done on a machine with an A6000 GPU, with Ubuntu 18.04, CUDA 11.7 and Pytorch 2.0.0. Float32 is used because numerical precision is key to our approach.

\subsection{Function Dataset}\label{dataset}

As described in Section \ref{randomfunc}, we get 9,024 polynomial functions from the \emph{parametric functions} dataset in AMPS dataset\cite{hendrycks21} that can be parsed by SymPy. Then we randomize each of them by randomly increase, decrease, multiply or divide each constant (including coefficient and exponent) by an integer between 1 and 3 without changing its sign. In this way we get 58,512 polynomial functions in total. 

Then we randomly generate trigonometric or exponential functions such as $a \cdot \sin(bx) + c$, $a \cdot \cos(bx) + c$ and $a \cdot e^{bx} + c$, where $a$, $b$ and $c$ are random constants between -10 and 10. 113,825 functions were generated in this way. 

Obviously we could have included more types of functions. But we feel the above set is sufficient for studying how deep learning can discover the way to calculate integrals.

\subsection{Symbolic Regression}\label{exp:symreg}

For each function $f$ in our function dataset, we compute the integral defined as area under the curve $g(x) = \int_{0}^{x} f(t) dt$. We use numerical methods to compute the value of $g(x)$ on the range $[0, T]$, using the Trapezoidal rule\footnote{Using function numpy.trapz.} which is simply a cumulative sum of a function over $[0, T]$ on the $x$-axis. We set $T=4$, and compute values on 10,000 points for each function.

This gives us 10,000 pairs of $(x, g(x))$ for each function. We apply the symbolic regression model \cite{dAscoli22} mentioned in Section \ref{symbolic}, which takes a number of $(x,y)$ pairs as input and outputs a function that is inferred by the underlying deep learning model, whose coefficients are adjusted by gradient descent to minimize the error against the input points. We denote the output function of symbolic regression as $g'(x)$.  

The symbolic regression model is not always accurate, and we need to filter out cases where $g'(x)$ is obviously incorrect. As discussed in the end of Section \ref{model}, we compute the average relative absolute difference between $g(x)$ and $g'(x)$ over the 10,000 points, and ignore a case if the difference is greater than $\epsilon=0.01$. After this filtering we have 49,109 polynomial functions left (out of 58,512), and 63,409 non-polynomial functions left (out of 113,825). Out of these 112.5K examples, We use 90K as our training set, and 22.5K as our evaluation set which is used to tune the hyper-parameters. We generated additional 10K polynomial and 10K non-polynomial functions in the same way as our testing dataset. 

When running symbolic regression, it is possible that given the data points of $cos(x)$, a symbolic regression model outputs $sin(\frac{\pi}{2}-x)$. It is also possible that given the data points of $a \cdot e^{bx}$, the model outputs $a' \cdot e^{bx-c}$, where $a' = a \cdot e^c$. Fortunately the symbolic regression approach we choose prefers simpler outputs, and thus in the majority of cases we do not observe the above issue. 

Please note that a  $g'(x)$ can still be incorrect even if its relative error is smaller than $\epsilon$. Since our goal is to use deep learning to learn to infer the integral of functions purely on its own (without human supervision), we have to live with the errors of symbolic regression. Later we will show that our models are more accurate than symbolic regression, and are much faster because only it only needs to infer a sequence of tokens.

%We did a manual spot check on these functions, and they are mostly correct, except that a small percentage of them have different forms. For example, $\sin(x)$ may become $\cos(x-1.57)$ after symbolic regression. Since the symbolic regression model prefers simpler math expressions, it is rare to observe such cases.

\subsection{Training Models for Integral}\label{exp:model}

As described above, we have acquired 112.5K examples, each having a function $f(x)$ and its integral $g(x) = \int_{0}^{x} f(t) dt$.  For example, $f(x)=2x^2 + 3.55x - 1.4$, and $g(x)=0.67x^3+1.78x^2-1.4x$ is the output of symbolic regression. Please note that the raw output of symbolic regression can have many digits (e.g., $1.77501292$) and we round all coefficients to two decimal places in our study.

Section \ref{model} described the two models we use in our study: An autoregressive model called GPT-Neo, and a sequence-to-sequence model called Flan-T5. We train both models to learn to predict the integral $g(x)$ by taking the original function $f(x)$ as input. For GPT-Neo, we need to convert each pair of $f(x)$ and $g(x)$ into a string, which will be split into tokens and fed into the GPT-Neo model to learn to predict the next token based on previous tokens. We use SymPy \cite{sympy} to convert a math expression into a string. For the above example, it is converted into 

``$2*x**2+3.55*x-1.4 \: entail \: 0.67*x**3+1.78*x**2-1.4*x$''

Please note we use a special token "entail" to separate the original function and its integral. At inference time, we can feed ``$2*x**2+3.55*x-1.4 \: entail$'' into the model, and it should output the integral, one token at a time.

Unlike GPT-Neo, Flan-T5 is a sequence-to-sequence model, and we simply use the original function (such as ``$2*x**2+3.55*x-1.4$'') as the input sequence, and the integral as the output sequence (such as ``$0.67*x**3+1.78*x**2-1.4*x$'').

\subsubsection{Accuracy Evaluation}\label{exp:accr}

Our models may not always produce the correct expression for integral. It might be due to the model's capability, or due to errors in its training data generated by symbolic regression. Given a function $f$ and its integral $g(x) = \int_{0}^{x} f(t) dt$ (rounded to two decimal places), suppose a model's output is $\hat{g}(x)$. We say $\hat{g}(x)$ is correct if $\hat{g}(x) = g(x)$. 

Sometimes $\hat{g}(x)$ only differs a little from $g(x)$, which is probably due to rounding errors. Consider our example mentioned above: $f(x)=2x^2 + 3.55x - 1.4$ and $g(x)=0.67x^3+1.78x^2-1.4x$. Our model may produce $\hat{g}(x)=0.67x^3+1.77x^2-1.4x$, since symbolic regression may produce such a result which is almost equally good as $g(x)$. We define the error between $g(x)$ and $\hat{g}(x)$ in the same way as we define the error of symbolic regression:

\begin{equation}\label{pred_diff}
	RelativeDiff(g(x), \hat{g}(x)) = \frac{\int_{0}^{T} |g(x) - \hat{g}(x)| dx}{T \cdot max(max(|g(x)|), max(|\hat{g}(x)|))}
\end{equation}

We say $\hat{g}(x)$ is approximately correct if $RelativeDiff(g(x), \hat{g}(x)) < \epsilon$, where $\epsilon=0.01$. In the above example, $\hat{g}(x)=0.67x^3+1.77x^2-1.4x$ is not correct but is approximately correct.

\subsubsection{Pre-training}\label{exp:pretrain}

Although there are open-source pre-trained version of each model available on the web, we do not use them because the pre-training may involve knowledge about calculus. Instead, we pre-train each model using the \emph{algebra} and \emph{number theory} datasets in the AMPS$\rightarrow$Mathematica dataset \cite{hendrycks21}, which contain no knowledge of calculus. These datasets contain 1.99M math problems and answers, mostly at grade-school level. The pre-training should teach each model about basic math knowledge on algebra and number theory.

Table \ref{tab:pretrain} shows how GPT-Neo's accuracy improves with pre-training. We can see that pre-training improves the percentage of correct from 74.1\% to 80.2\%, and that of approximately correct from 85.9\% to 95.3\%. The accuracy is about 2\% lower than that with the open-source model, probably because the latter has been trained with some knowledge about calculus.

\begin{table}[h]
	\caption{GPT-Neo's Performances with various amounts of pre-training. In each case the model is first trained with pre-trainined data, and then trained five epochs on the our dataset for inferring integral of functions}\label{tab:pretrain}%
	\begin{tabular}{@{}llll@{}}
		\toprule
		Pre-Training &  \%-Correct & \%-Approx.-Correct \\
		\midrule
		Randomly Initialized    & 74.1\%  & 85.9\%    \\
		After 2 epochs   & 78.4\%  & 93.0\%  \\
		After 5 epochs   & 80.2\%  & 95.2\%   \\
		After 8 epochs   & 80.2\%  & 95.3\%   \\
		From HuggingFace\footnotemark[1]  & 82.4\%  & 97.3\%    \\
		\botrule
	\end{tabular}
	\footnotetext[1]{Downloaded from https://huggingface.co/xhyi/PT\_GPTNEO350\_AT}
\end{table}

\subsubsection{Model Accuracy}\label{exp:modelaccr}

In this section we report the accuracy of our models in inferring integral of a function. As mentioned in Section  \ref{exp:symreg}, we have a training set of 90K examples, a validation set of 22.5K, and a testing set of 20K (half polynomials and half non-polynomials). We train our models using the training set, based on the pre-trained model. We use the validation set to tune the hyper-parameters, and we report accuracy on the testing set. 

For Flan-T5, we found the validation loss stabilized in the fifth epoch, and thus we trained the model for five epochs. We used learning rate of $3 \times 10^{-4}$ and batch size of 8. Figure 3 shows its accuracy on inferring integrals of polynomial functions, and how it changes w.r.t. number of examples trained (90K $\times$ 5 epochs = 450K examples in total). We can see that the \%-Correct grows from 46.2\% to 67.7\%, and \%-Approx.-Correct grows from 64.0\% to 85.7\%. 

Figure 4 shows Flan-T5's accuracy on inferring integrals of non-polynomial functions. The \%-Correct grows from 45.1\% to 64.6\%, and \%-Approx.-Correct grows from 52.2\% to 75.7\%. 

We can see Flan-T5 is not highly accurate, and we find that GPT-Neo has much higher accuracy, as shown below.

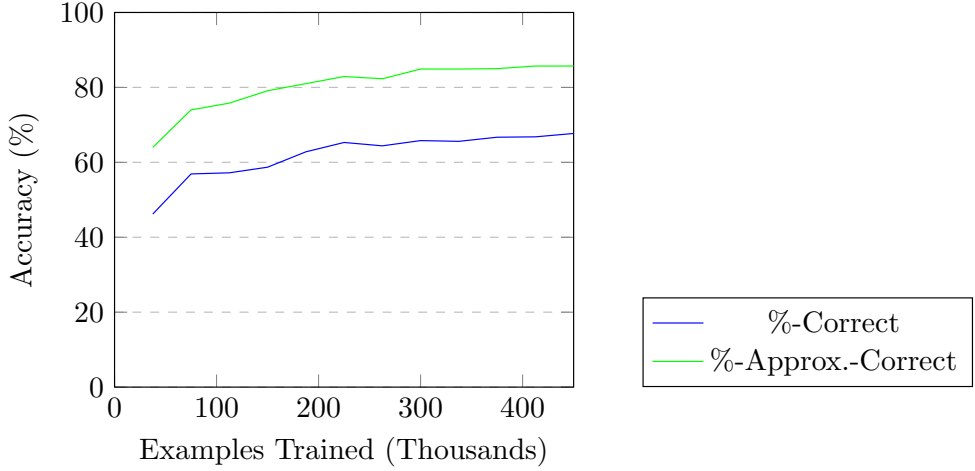
\begin{figure}
	\begin{center}
		\begin{minipage}{1.0\textwidth}
			\centering
			\resizebox{\columnwidth}{!}{
				\begin{tikzpicture}
					\begin{axis}[
						title={},
						xlabel={Examples Trained (Thousands)},
						ylabel={Accuracy (\%)},
						xmin=0, xmax=450,
						ymin=0, ymax=100,
						xtick={0,100,200,300,400},
						ytick={0,20,40,60,80,100},
						legend style={at={(1.15,0)},anchor=south west},
						ymajorgrids=true,
						grid style=dashed,
						]
						\addplot[
						color=blue,
						mark=circle,
						]
						coordinates {
							(37.5, 46.2)
							(75, 56.9)
							(112.5, 57.2)
							(150, 58.7)
							(187.5, 62.8)
							(225, 65.3)
							(262.5, 64.4)
							(300, 65.8)
							(337.5, 65.6)
							(375, 66.7)
							(412.5, 66.8)
							(450, 67.7)
						};
						\addlegendentry{\%-Correct}
						\addplot[
						color=green,
						mark=circle,
						]
						coordinates {
							(37.5, 64)
							(75, 74)
							(112.5, 75.8)
							(150, 79.1)
							(187.5, 81.0)
							(225, 82.9)
							(262.5, 82.3)
							(300, 84.9)
							(337.5, 84.9)
							(375, 85.0)
							(412.5, 85.7)
							(450, 85.7)
						};
						\addlegendentry{\%-Approx.-Correct}
					\end{axis}
				\end{tikzpicture}
			}
		\end{minipage}
	\end{center}
	\caption{The \%-Correct and \%-Approximately-Correct of Flan-T5 in inferring integrals of polynomial functions.}
\end{figure}

\begin{figure}\label{fig:flan-nonpoly}
	\begin{center}
		\begin{minipage}{1.0\textwidth}
			\centering
			\resizebox{\columnwidth}{!}{
				\begin{tikzpicture}
					\begin{axis}[
						title={},
						xlabel={Examples Trained (Thousands)},
						ylabel={Accuracy (\%)},
						xmin=0, xmax=450,
						ymin=0, ymax=100,
						xtick={0,100,200,300,400},
						ytick={0,20,40,60,80,100},
						legend style={at={(1.15,0)},anchor=south west},
						ymajorgrids=true,
						grid style=dashed,
						]
						\addplot[
						color=blue,
						mark=circle,
						]
						coordinates {
							(37.5, 45.1)
							(75, 50.8)
							(112.5, 54.6)
							(150, 58.4)
							(187.5, 60.3)
							(225, 62.7)
							(262.5, 62.7)
							(300, 64.3)
							(337.5, 62.3)
							(375, 64.4)
							(412.5, 63.7)
							(450, 64.6)
						};
						\addlegendentry{\%-Correct}
						\addplot[
						color=green,
						mark=circle,
						]
						coordinates {
							(37.5, 52.2)
							(75, 60.9)
							(112.5, 63.8)
							(150, 68.4)
							(187.5, 70.6)
							(225, 72.8)
							(262.5, 73.3)
							(300, 74.6)
							(337.5, 73.8)
							(375, 74.7)
							(412.5, 74.2)
							(450, 75.7)
						};
						\addlegendentry{\%-Approx.-Correct}
					\end{axis}
				\end{tikzpicture}
			}
		\end{minipage}
	\end{center}
	\caption{The \%-Correct and \%-Approximately-Correct of Flan-T5 in inferring integrals of non-polynomial functions.}
\end{figure}
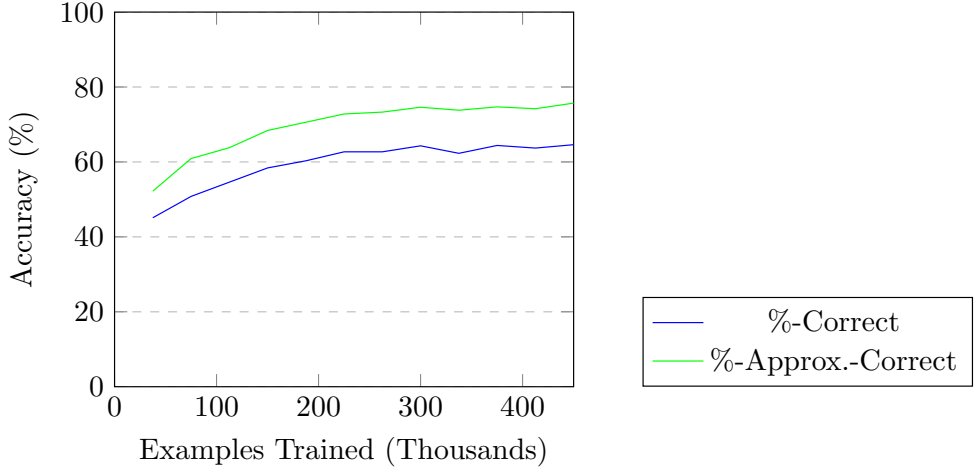

For GPT-Neo, we also trained the model for five epochs, using learning rate of $2 \times 10^{-5}$ and batch size of 16. For polynomials, \%-Correct improves from 57.0\% to 73.3\% over the course of training, and \%-Approximately-Correct (i.e., no more than 1\% average relative error, which is probably due to rounding errors in training data caused by symbolic regression) improves from 77.6\% to 95.7\%. 

For non-polynomials,  \%-Correct improves from 67.4\% to 87.1\% over the course of training, and \%-Approximately-Correct improves from 74.9\% to 94.8\%. 

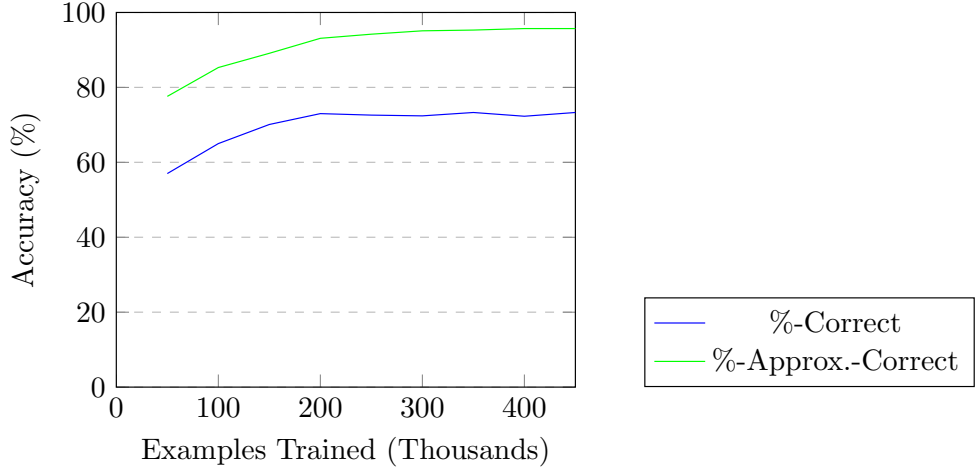
\begin{figure}\label{fig:gpt-poly}
	\begin{center}
	\begin{minipage}{1.0\textwidth}
		\centering
		\resizebox{\columnwidth}{!}{
			\begin{tikzpicture}
				\begin{axis}[
					title={},
					xlabel={Examples Trained (Thousands)},
					ylabel={Accuracy (\%)},
					xmin=0, xmax=450,
					ymin=0, ymax=100,
					xtick={0,100,200,300,400},
					ytick={0,20,40,60,80,100},
					legend style={at={(1.15,0)},anchor=south west},
					ymajorgrids=true,
					grid style=dashed,
					]
					\addplot[
					color=blue,
					mark=circle,
					]
					coordinates {
						(50, 57.0)
						(100, 65.0)
						(150, 70.1)
						(200, 73.0)
						(250, 72.6)
						(300,72.4)
						(350,73.3)
						(400,72.3)
						(450,73.3)
					};
					\addlegendentry{\%-Correct}
					\addplot[
					color=green,
					mark=circle,
					]
					coordinates {
						(50, 77.6)
						(100, 85.3)
						(150, 89.1)
						(200, 93.1)
						(250, 94.2)
						(300,95.1)
						(350,95.3)
						(400,95.7)
						(450,95.7)
					};
					\addlegendentry{\%-Approx.-Correct}
				\end{axis}
			\end{tikzpicture}
		}
	\end{minipage}
	\end{center}
	\caption{The \%-Correct and \%-Approximately-Correct of GPT-Neo in inferring integrals of polynomial functions.}
\end{figure}

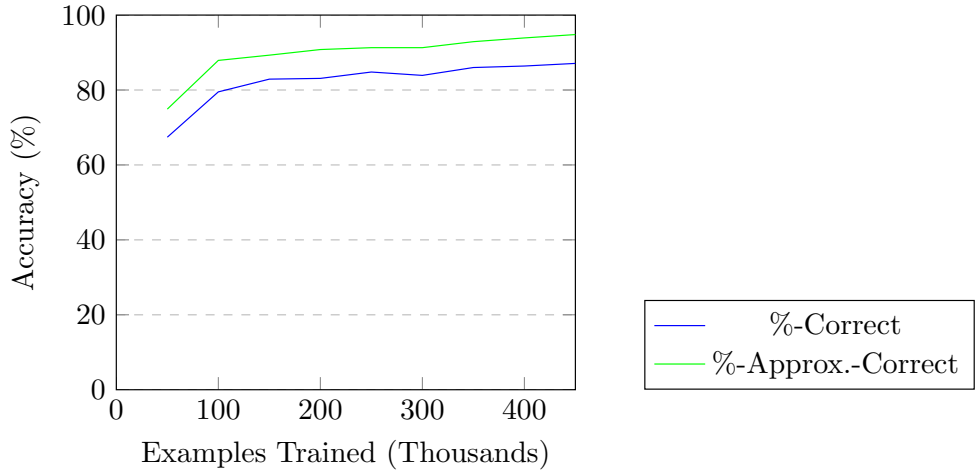
\begin{figure}\label{fig:gpt-nonpoly}
	\begin{center}
		\begin{minipage}{1.0\textwidth}
			\centering
			\resizebox{\columnwidth}{!}{
				\begin{tikzpicture}
					\begin{axis}[
						title={},
						xlabel={Examples Trained (Thousands)},
						ylabel={Accuracy (\%)},
						xmin=0, xmax=450,
						ymin=0, ymax=100,
						xtick={0,100,200,300,400},
						ytick={0,20,40,60,80,100},
						legend style={at={(1.15,0)},anchor=south west},
						ymajorgrids=true,
						grid style=dashed,
						]
						\addplot[
						color=blue,
						mark=circle,
						]
						coordinates {
							(50, 67.4)
							(100, 79.5)
							(150, 82.9)
							(200, 83.1)
							(250, 84.8)
							(300,83.9)
							(350,86.0)
							(400,86.4)
							(450,87.1)
						};
						\addlegendentry{\%-Correct}
						\addplot[
						color=green,
						mark=circle,
						]
						coordinates {
							(50, 74.9)
							(100, 87.9)
							(150, 89.3)
							(200, 90.8)
							(250, 91.3)
							(300,91.3)
							(350,92.9)
							(400,93.9)
							(450,94.8)
						};
						\addlegendentry{\%-Approx.-Correct}
					\end{axis}
				\end{tikzpicture}
			}
		\end{minipage}
	\end{center}
	\caption{The \%-Correct and \%-Approximately-Correct of GPT-Neo in inferring integrals of non-polynomial functions.}
\end{figure}

\begin{table}[h]
	\caption{Comparing Performances of GPT-Neo, Flan-T5, and Symbolic Regression in computing integrals}\label{tab:compare}%
	\begin{tabular}{@{}llll@{}}
		\toprule
		  & GPT-Neo  & Flan-T5 & Symbolic Regression \\
		\midrule
		\%-Correct Polynomials   & 73.3\%  & 67.7\%  & 65.7\%  \\
		\%-Approx-Correct Polynomials    & 95.7\%   & 85.7\%  & 90.6\%  \\
		\%-Correct Non-polynomials   & 87.1\%   & 64.6\%  & 46.4\%  \\
		\%-Approx-Correct Non-polynomials    & 94.8\%   & 75.7\%  & 71.1\%  \\
		Avg. Time (sec)    & 0.722   & 0.151  & 3.17  \\
		\botrule
	\end{tabular}
\end{table}

Table \ref{tab:compare} compares the performances of GPT-Neo, Flan-T5, and the Symbolic Regression model in \cite{dAscoli22}. Please note that GPT-Neo and Flan-T5 directly infer the integral function from the original function, without any numeric computation. In contrast, in order to use the Symbolic Regression model to infer the integral of a function, we need to first use the Trapezoidal rule to compute the values of the integral function, and then use the model to infer the math expression of the function. 

From Table \ref{tab:compare} we can see that GPT-Neo is the most accurate among the three. As mentioned before, the training data of GPT-Neo and Flan-T5 contain some rounding errors, because the training data was generated using symbolic regression which is not 100\% precise. Such small errors in the training data leads to some small errors (usually $<$1\%) in the output integral functions of GPT-Neo and Flan-T5. We can see that although GPT-Neo produces exactly the expected integral function in about 80\% of cases, it has 95\%+ accuracy if we allow such small errors ($<$1\%).

\subsection{Finding Rules for Integrals}

\subsubsection{Rules of integral for polynomials}

As described in Section \ref{discover-rule}, we aim at finding very simple rules for inferring of integrals of each type of functions. For example, if we consider quadratic functions as a type, the integral of function $f(x)=a_2 x^2 +a_1 x + a_0$ is $g(x) = a_3' x^3 + a_2' x^2 + a_1' x + a_0$, the basic rules would be $a_3' =\frac{a_2}{3}$, $a_2' = \frac{a_1}{2}x^2$, $a_1' = a_0$, and $a_0'=0$\footnote{$a_0'=0$ because we define integrals as area under the curve starting from $x=0$.}. 

Please note that we compute the integral of a function by first applying Trapezoidal rule to compute the integral function's values, and then using the symbolic regression model in \cite{dAscoli22} to find the math expression, and excluding all cases where the average relative error is greater than 1\%. Therefore, the discovery of rules of integrals does not depend on our ability to train models to infer integral functions.

In this study, we design a \emph{rule search system}, which tries every possible combination of single, double (and more) coefficients, combined together with arithmetic operators, and checks if that can be used to infer a coefficient in the integral function  $g(x)$. In order to find a rule to infer a particular coefficient $a_k'$ in $g(x)$, the \emph{rule search system} starts from trying each individual coefficient $a_i$ (and also $\frac{1}{a_i}$) in $f(x)$ by running linear regression with $a_i$ and $a_k'$. If it finds an $a_i$ that leads to a high $R^2$ in the linear regression ($|R^2| > 0.95$), it stops and declares success. Otherwise it tries each pair of coefficients $a_i$ and $a_j$ in $f(x)$, by combining them using each arithmetic operation. If no success with pairs of coefficients, it tries triples, and stops there.

\begin{table}[h]
	\caption{$R^2$ of linear regression between individual coefficients of $f(x)$ ($a_0, ..., a_6$) and those of $g(x)$  ($a_1', ..., a_7'$), for polynomial functions }\label{tab:poly_lr}%
	\begin{tabular}{@{}llllllll@{}}
		\toprule
		$R^2$ & $a_0$  & $a_1$ & $a_2$ & $a_3$  & $a_4$ & $a_5$ & $a_6$ \\
		\midrule
		$a_1'$  & \bf{1.000}  &  0.006  & 0.177  & 0.036  & 0.095  & -0.021  & -0.031 \\
		$a_2'$  & 0.006 & \bf{1.000}  & 0.002  & 0.137  &  0.010 & 0.005  & -0.005 \\
		$a_3'$  & 0.176  &  -0.001 &  \bf{1.000}  & -0.045  &  -0.115 & -0.027  & -0.043 \\
		$a_4'$  & 0.036  & 0.142  & -0.033  & \bf{1.000}  &  -0.016 &  -0.006 & -0.010 \\
		$a_5'$  & 0.095  &  0.009 & -0.114  & -0.023  & \bf{1.000}  & -0.015  & -0.023 \\
		$a_6'$  & -0.020  &  0.005 & -0.027  & -0.006  &  -0.015 & \bf{1.000}  & -0.020 \\
		$a_7'$  & -0.031  & -0.002  & 0.039  &  -0.006 & -0.022  &  0.020 & \bf{1.000} \\
		\botrule
	\end{tabular}
	\footnotetext{Note: $a_0' \approx 0$  in almost all cases and thus does not require further studies.}
\end{table}

Table \ref{tab:poly_lr} shows the $R^2$ of linear regression between individual coefficients of $f(x)$ and those of $g(x) = \int_{0}^{x} f(t) dt$, where $f(x)$ is a polynomial function and $a_i$ is the coefficient for $x^i$. We can see that each coefficient of $g(x)$ ($a_1', ..., a_7'$) has $R^2$ of 1.0 with a particular coefficient of $f(x)$ ($a_0, ..., a_6$). To be more specific, $a_k'$ has perfect correlation with $a_{k-1}$, and has little correlation with other coefficients of $f(x)$. This shows that each $a_k'$ can be perfectly predicted by $a_{k-1}$.

\begin{table}[h]
	\caption{Slope and intercept of linear regression between each coefficient $a_k'$ of $g(x)$ and the corresponding coefficient in $f(x)$ that is correlated with $a_k'$ (i.e. $a_{k-1}$), for polynomial functions }\label{tab:poly_slope}%
	\begin{tabular}{@{}lllllll@{}}
		\toprule
		 & Slope  & Intercept &  & & & \\
		\midrule
		$a_1'$  & 1.000  &  -0.003 & &  &  & \\
		$a_2'$  & 0.500 & -0.002 & &  &  &  \\
		$a_3'$  & 0.333  &  -0.003 & &  &  & \\
		$a_4'$  & 0.250  & 0.000  & &  &  & \\
		$a_5'$  & 0.200  &  0.000 & &  &  &  \\
		$a_6'$  & 0.167  &  0.000 & &  &  &  \\
		$a_7'$  & 0.143  & 0.000 & &  &  & \\
		\botrule
	\end{tabular}
\end{table}

Table \ref{tab:poly_slope} shows the slope and intercept of the linear regression between each coefficient $a_k'$ of $g(x)$ and the corresponding coefficient in $f(x)$ that is correlated with $a_k'$ (i.e., $|R^2|$ is close to 1.0). We can see that the intercept is close to zero, and therefore the formula for integrating polynomials (up to 6th power) can be written as 

\begin{equation}
	\begin{split}
	\int_{0}^{x} a_0 + a_1t+a_2t^2 + a_3t^3 + a_4t^4 +a_5t^5 + a_6t^6 dt = \\ a_0t + 0.5 a_1t^2 + 0.333 a_2t^3 + 0.25 a_3t^4 + 0.2 a_4t^5 + 0.167 a_5t^6 + 0.143 a_6t^7 
	\end{split}
\end{equation}

Now we let our rule search system find the connection between the slope $\beta$ and the power $p$ (e.g., $a_i'$ is associated with power $i$), because they are the only two meaningful variables left. It first runs linear regression between $p$ and $\beta$, and gets $R^2$ of 0.735, which is not high enough to confirm their connection. Then it tries to find the relationship between $\beta$ and a simple transform of $p$ using any arithmetic operator, and finds that $\frac{1}{p}$ has perfect correlation ($R^2 = 1.00$) with $\beta$, and $\beta = 1.00 \cdot \frac{1}{p} + 0.001$. Our rule search system discards any constants smaller than 0.01 because they are likely caused by rounding errors. Thus it concludes that $\beta = \frac{1}{p}$.

Combining with the results in \ref{tab:poly_lr}, it concludes that the coefficients of $g(x)$ can be inferred as 

\begin{equation}
a_k' = \frac{1}{k} a_{k-1},
\end{equation}

which means that 

\begin{equation}
	\int_{0}^{x} a_{k-1} t^{k-1} dt = \frac{a_{k-1}}{k} x^k
\end{equation}

\subsubsection{Rules of integral for exp function}

Now let us look at non-polynomial functions. Since we studied three types of non-polynomial functions ($exp$, $sin$ and $cos$), we need to study their integral rules separately. 

We start by applying our \emph{rule search system} on the 15.3K cases involving $exp$ functions, which take the form of $a_4 exp(a_3x+a_2) + a_1x +a_0$, where $a_2$ and $a_1$ are both zero in our generated functions. The \emph{rule search system} found that among their integral functions, 99.9\% of them take the form of $a_4' exp(a_3'x+a_2') + a_1'x +a_0'$, and concluded that this is the form of $exp$ function's integral. It was also found that $a_2'$ are mostly zero and thus require no further investigation. The correlation (indicated by $R^2$ of linear regression) of the remaining coefficients is shown in Table \ref{tab:lr_exp}. 

\begin{table}[h]
	\caption{$R^2$ of linear regression between individual coefficients of $f(x)$ ($a_0, ..., a_4$) and those of the integral function $g(x)$  ($a_0', ..., a_4'$) , where $f(x)=a_4 exp(a_3x+a_2) + a_1x +a_0$}\label{tab:lr_exp}%
	\begin{tabular}{@{}llllll@{}}
		\toprule
		$R^2$ & $a_0$  & $a_1$ & $a_2$ & $a_3$  & $a_4$  \\
		\midrule
		$a_0'$  & 0.121 & N/A  & N/A\footnotemark[1]  & -0.067  &  0.707  \\
		$a_1'$  & \bf{1.000} & N/A  & N/A  & -0.027  &  0.133  \\
		$a_3'$  & -0.027  & N/A  & N/A & \bf{1.000}  &  -0.013  \\
		$a_4'$  & -0.121  &  N/A & N/A  & 0.069  & -0.706   \\
		\botrule
	\end{tabular}
	\footnotetext{Note: $a_2'=0$ in most cases and thus does not require further studies.}
	\footnotetext[1]{$a_1$ and $a_2$ are all zeros and linear regression could not be performed.}
\end{table}

From Table \ref{tab:lr_exp} one can see that $a_1'$ is perfectly correlated with $a_0$ and $a_3'$ perfectly correlated with $a_3$, which are consistent with our expectation. Based on the results of linear regressions, it is found that $a_1' \approx a_0$ (slope 1.000 and intercept 0.000), and $a_3' \approx a_3$ (slope 1.000 and intercept 0.000). But there is no conclusion drawn for $a_0'$ and $a_4'$, and the \emph{rule search system}  investigates pairs of coefficients among $a_0, ..., a_4$, combined with an arithmetic operation.

\begin{table}[h]
	\caption{The maximum $R^2$ of linear regression between $a_4'$ (of integral of a function involving $exp$) and any pair of coefficients of the original function, combined by an arithmetic operator. We also show the slope and intercept of linear regression. }\label{tab:exp_pair}%
	\begin{tabular}{@{}llllll@{}}
		\toprule
		Operator & $R^2$ with max abs. value & Slope  & Intercept  & First Coefficient & Second Coefficient \\
		\midrule
		Addition\footnotemark[1]  & -0.706\footnotemark[2]  &  -0.124 & -0.013 & $a_4$ & $a_4$  \\
		Multiplication  & 0.807 & 0.080 & -0.033 & $a_3$  & $a_4$   \\
		Division  & \bf{1.000} &  1.000 & 0.001 & $a_4$ & $a_3$ \\
		\botrule
	\end{tabular}
	\footnotetext[1]{A multi-variate linear regression is performed to use the pairs of coefficients to predict $a_4'$, which allows the pair of coefficients to be weighted summed.}
	\footnotetext[2]{A linear regression using $a_0, ..., a_4$ and $a_4'$ also has $|R^2|$ of 0.706.}
\end{table}

From Table \ref{tab:exp_pair} it is found that the only combination of coefficients leading to qualified result ($|R^2| > 0.95$) is $\frac{a_4}{a_3}$. The formula found is $a_4' = 1.000 \frac{a_4}{a_3} + 0.001$, and we can confidently draw the conclusion of  $a_4' = \frac{a_4}{a_3}$\footnote{Any constant smaller than 0.01 is discarded because it is probably caused by rounding errors.}.  

\begin{table}[h]
	\caption{The maximum $R^2$ of linear regression between $a_0'$ (of integral of a function involving $exp$) and any pair of coefficients of the original function, combined by an arithmetic operator. We also show the slope and intercept of linear regression. }\label{tab:exp_pair_a0}%
	\begin{tabular}{@{}llllll@{}}
		\toprule
		Operator & $R^2$ with max abs. value & Slope  & Intercept  & First Coefficient & Second Coefficient \\
		\midrule
		Addition\footnotemark[1]  & -0.706  &  -0.124 & -0.013 & $a_4$ & $a_4$  \\
		Multiplication  & -0.807 & -0.080 & 0.034 & $a_3$  & $a_4$   \\
		Division  & \bf{-0.999} &  -0.999 & 0.000 & $a_4$ & $a_3$ \\
		\botrule
	\end{tabular}
		\footnotetext[1]{A multi-variate linear regression is performed to use the pairs of coefficients to predict $a_4'$, which allows the pair of coefficients to be weighted summed.}
\end{table}

Similarly we find out that $a_0' = \frac{a_4}{a_3}$, with results in Table \ref{tab:exp_pair_a0}. Combining with previous results, our \emph{rule search system} finds that for $f(x) = a_4 exp(a_3x) +a_0$, its integral is $\int_{0}^{x} f(t) dt = \frac{a_4}{a_3} exp(a_3x) + a_0x - \frac{a_4}{a_3}$.

\subsubsection{Rules of integral for sin function}

Then we apply our \emph{rule search system} on the 23.6K cases involving $sin$ functions, which take the form of $a_4 sin(a_3x+a_2) + a_1x +a_0$, where $a_2$ and $a_1$ are both zero in our generated functions. The \emph{rule search system} found that among their integral functions, 93.8\% of them take the form of $a_4' cos(a_3'x+a_2') + a_1'x +a_0'$, and concluded that this is the form of $sin$ function's integral. It was also found that $a_2'$ are mostly zero and thus require no further investigation. The correlation (indicated by $R^2$ of linear regression) of the remaining coefficients is shown in Table \ref{tab:poly_lr_sin}. 

\begin{table}[h]
	\caption{$R^2$ of linear regression between individual coefficients of $f(x)$ ($a_0, ..., a_4$) and those of the integral function $g(x)$  ($a_0', ..., a_4'$), where $f(x) = a_4 sin(a_3x) + +a_0$ }\label{tab:poly_lr_sin}%
	\begin{tabular}{@{}llllll@{}}
		\toprule
		$R^2$ & $a_0$  & $a_1$ & $a_2$ & $a_3$  & $a_4$  \\
		\midrule
		$a_0'$  & 0.070 & N/A  & N/A\footnotemark[1]  & -0.036  &  0.900  \\
		$a_1'$  & \bf{1.000} & N/A  & N/A  & 0.056  &  0.073  \\
		$a_3'$  & 0.058  & N/A  & N/A & \bf{0.995}  &  -0.012  \\
		$a_4'$  & -0.062  &  N/A & N/A  & 0.032  & -0.885   \\
		\botrule
	\end{tabular}
	\footnotetext{Note: $a_2'=0$ in most cases and thus do not require further studies.}
	\footnotetext[1]{$a_1$ and $a_2$ are all zeros and linear regression could not be performed.}
\end{table}

From Table \ref{tab:poly_lr_sin} we can see that $a_1'$ is perfectly correlated with $a_0$, and the same between $a_3'$ and $a_3$, which are consistent with our expectations. Based on the results of linear regressions, it is found that $a_1' = a_0$ (slope 1.000 and intercept 0.000), and $a_3' \approx a_3$ (slope 0.999 and intercept -0.001). Again there is no conclusion for $a_0'$ and $a_4'$, which has some correlation with $a_4$ but is lower than our requirement of $|R^2| > 0.95$. Thus the \emph{rule search system} needs to explore pairs of coefficients combined with arithmetic operators.

\begin{table}[h]
	\caption{The maximum $R^2$ of linear regression between $a_4'$ (of integral of $sin$ function) and any pair of coefficients of $f(x)$, combined by each operator. We also show the slope and intercept of linear regression. }\label{tab:sin_pair}%
	\begin{tabular}{@{}llllll@{}}
		\toprule
		Operator & $R^2$ with max abs. value & Slope  & Intercept  & First Coefficient & Second Coefficient \\
		\midrule
		Addition\footnotemark[1]  & -0.894  &  -0.276 & 0.037 & $a_4$ & $a_3$  \\
		Multiplication  & -0.763 & -0.052 & 0.012 & $a_3$  & $a_4$   \\
		Division  & \bf{0.997} &  0.997 & -0.001 & $a_4$ & $a_3$ \\
		\botrule
	\end{tabular}
	\footnotetext[1]{A multi-variate linear regression is performed to use the pairs of coefficients to predict $a_4'$, which allows the pair of coefficients to be weighted summed.}
\end{table}

From Table \ref{tab:sin_pair} it is found that the only combination of coefficients leading to qualified result ($|R^2| > 0.95$) is $\frac{a_4}{a_3}$. The formula found is $a_4' = 0.997 \frac{a_4}{a_3} - 0.001$. Since anything less than 0.01 is likely to be caused by rounding errors, we can draw the conclusion of  $a_4' =  \frac{a_4}{a_3}$. 

\begin{table}[h]
	\caption{The maximum $R^2$ of linear regression between $a_0'$ (of integral of $sin$ function) and any pair of coefficients of $f(x)$, combined by each operator. We also show the slope and intercept of linear regression. }\label{tab:sin_pair_a0}%
	\begin{tabular}{@{}llllll@{}}
		\toprule
		Operator & $R^2$ with max abs. value & Slope  & Intercept  & First Coefficient & Second Coefficient \\
		\midrule
		Addition\footnotemark[1]  & 0.900  &  0.292 & 0.016 & $a_4$ & $a_3$  \\
		Multiplication  & 0.783 & 0.065 & -0.007 & $a_3$  & $a_4$   \\
		Division  & \bf{0.996} &  0.998 & -0.001 & $a_4$ & $a_3$ \\
		\botrule
	\end{tabular}
		\footnotetext[1]{A multi-variate linear regression is performed to use the pairs of coefficients to predict $a_4'$, which allows the pair of coefficients to be weighted summed.}
\end{table}

Similarly the \emph{rule search system} finds that $a_0' = -\frac{a_4}{a_3}$, with results in Table \ref{tab:sin_pair_a0}. Combining previous results, it finds that for $f(x) = a_4 sin(a_3x) +a_0$, its integral is $\int_{0}^{x} f(t) dt = -\frac{a_4}{a_3} cos(a_3x) + a_0x + \frac{a_4}{a_3}$.

Integrating functions involving $cos$ is similar to that for $sin$, and the \emph{rule search system} concludes that for $f(x) = a_4 cos(a_3x) +a_0$, its integral is $\int_{0}^{x} f(t) dt = \frac{a_4}{a_3} sin(a_3x) + a_0x$.

\section{Conclusions and Future Work}

The recent advances in large language models have shown that AI can solve difficult math problems or write code for programming challenges. However, being able to solve a type of well-defined problems after being trained on a huge corpus is very different from making discoveries in scientific research.

In this paper we explore how to use deep learning models to discover an old but most important math tool: Integrals. We show that with a definition of integrals (defined as area under the curve in our study), AI models can learn to infer the integral of a function, such as inferring that $\int_{0}^{x} t^2 dt = \frac{x^3}{3}$ and $\int_{0}^{x} ae^{bt} dt = \frac{a}{b} e^{bx} - \frac{a}{b}$. Our experiments show that deep learning models can learn to infer the integral function using autogressive or sequence-to-sequence models (similar to generating text or translating from one language to another one), with high accuracy. We also designed a \emph{rule search system} which can automatically discover the basic rules of integrals such as $\int_{0}^{x} t^n dt = \frac{x^{n+1}}{n+1}$.

Discovering how to compute integrals is just the first step of our journey towards automated research. In future we will explore how to let deep learning models learn to solve more sophisticated problems on their own, and the ultimate goal is to enable AI to assist human in making real scientific discoveries.


\bibliography{sn-bibliography}% common bib file

\end{document}